\documentclass[lettersize,journal]{IEEEtran}
\usepackage{amsmath,amssymb,amsfonts}
\usepackage{stmaryrd}
\usepackage[super]{nth}
\usepackage{booktabs}

\usepackage{algorithmic}
\usepackage{algorithm}
\usepackage{array}
\usepackage[caption=false,font=normalsize,labelfont=sf,textfont=sf]{subfig}
\usepackage{textcomp}
\usepackage{stfloats}
\usepackage{hyperref}
\usepackage{verbatim}
\usepackage{graphicx,tabularx}
\usepackage{multirow}
\usepackage{cite}
\usepackage[switch]{lineno}
\usepackage{subcaption}
\usepackage{lineno}
\usepackage{array}
\usepackage{comment}
\usepackage[dvipsnames]{xcolor}

\definecolor{mygreen}{rgb}{0,0.5,0}

\begin{document}

\title{SAR Strikes Back: A New Hope for RSVQA}

%\title{Integrating remote sensing into demographic analysis: a methodology for linking environment to health data}

\author{Lucrezia Tosato, Sylvain Lobry~\IEEEmembership{IEEE member}, Flora Weissgerber, Laurent Wendling
        % <-this % stops a space
\thanks{This work is supported by \textit{Agence Nationale de la Recherche} (ANR) under the ANR-21-CE23-0011 project. The experiments conducted in this study were performed using HPC/AI resources provided by GENCI-IDRIS (Grant 2023-AD011012735R2)}% <-this % stops a space
\thanks{Lucrezia Tosato, Sylvain Lobry and Laurent Wendling are affiliated with LIPADE, Université Paris Cité, 75006 Paris, France. (e-mail: lucrezia.tosato@u-paris.fr, sylvain.lobry@u-paris.fr, laurent.wendling@u-paris.fr; Corresponding Author: Lucrezia Tosato)}
\thanks{Flora Weissgerber and Lucrezia Tosato are affiliated with the French Aerospace Lab, ONERA, Palaiseau, France. (e-mail: lucrezia.tosato@onera.fr, flora.weissgerber@onera.fr}
%\thanks{Manuscript received April 19, 2021; revised August 16, 2021.}
}

% The paper headers
\markboth{IEEE Journal of Selected Topics in Applied Earth
Observations and Remote Sensing,~Vol.~XX, No.~XX, August~YYYY}%
{Shell \MakeLowercase{\textit{et al.}}: A Sample Article Using IEEEtran.cls for IEEE Journals}

%\IEEEpubid{0000--0000/00\$00.00~\copyright~2021 IEEE}
% Remember, if you use this you must call \IEEEpubidadjcol in the second
% column for its text to clear the IEEEpubid mark.

\maketitle

\begin{abstract}
%% Text of abstract
Remote Sensing Visual Question Answering (RSVQA) is a task that automatically extracts information from satellite images. It then processes a question to predict the answer from the images in textual form, helping with the interpretation of the image. 

While different methods have been proposed to extract information from optical images with different spectral bands and resolutions, only recently have some preliminary studies started exploring very high-resolution Synthetic Aperture Radar (SAR) data. These studies leverage SAR's ability to capture electromagnetic information and operate in all atmospheric conditions. However, no research has compared the results obtained using SAR and optical imagery or explored methods to fuse the two modalities effectively.

This work investigates the integration of SAR images into the RSVQA task exploring the most effective way to combine them with optical images. %\textcolor{orange}{(Reviewer 2, comment 1)}

In our research, we carry out a study on different pipelines for the task of RSVQA taking into account information from both SAR and optical data. To this purpose, we also present a dataset that allows for the introduction of SAR images in the RSVQA framework.

We study two different pipelines for RSVQA to include SAR modality and introduce a dataset enabling SAR-based RSVQA. The first model is an End-to-End approach while the second is a two-stage framework. In the latter, relevant information is extracted from SAR, before being translated into natural language to be used in the second step which only relies on a language model to provide the answer.

Our results show that the second pipeline achieves strong performance using SAR alone, yielding an improvement of nearly 10\% in overall accuracy compared to the first one. We then explore various types of fusion methods to use SAR and optical images together. A fusion at the decision level achieves the best results on the proposed dataset, with a final F1-micro score of 75.00\% and an F1-average of 81.21\% for classification, as well as an overall accuracy of 75.49\% for VQA. We show that SAR data offers additional information when fused with the optical modality, particularly for questions related to specific land cover classes, such as water areas.
%\let\thefootnote\relax\footnote{This work is supported by \textit{Agence Nationale de la Recherche} (ANR) under the ANR-21-CE23-0011 project.\\The experiments conducted in this study were performed using HPC/AI resources provided by GENCI-IDRIS (Grant 2023-AD011012735R2).}
%This is very vague, I would remove at this point
%However, we also highlight issues and consequences that can arise when working with very unbalanced datasets.
\end{abstract}

%%Graphical abstract
%\begin{graphicalabstract}
%\includegraphics{grabs}
%\end{graphicalabstract}

%%Research highlights
\begin{comment}
\begin{highlights}
\item We introduce a new large-scale dataset for Visual Question Answering on fully open-access Synthetic Aperture Radar and Optical data.
\item We propose two pipelines to take into account Synthetic Aperture Radar for the task.
\item We show that using both modalities allows to obtain better performances.
\end{highlights}
\end{comment}
%% Keywords

%% Use \section commands to start a section

\section{Introduction}\label{sec:intro}
\sloppy
Public and private sector initiatives are facilitating access to a wide range of remote sensing images. A well-known example is the Sentinel satellite constellation launched in 2014 as part of the European Union's Copernicus programme. 
The mission provides free access to a wide range of imagery, including optical and Synthetic Aperture Radar (SAR) images. The optical images acquired by Sentinel-2 deliver high-resolution information based on the reflected sunlight.
In contrast, SAR images, captured with the Sentinel-1 satellites, use radar signals which penetrate clouds and operate effectively at night. They provide data on surface roughness, moisture content, and other physical properties through radar backscatter. Sentinel-1 images contain two polarization channels, VV and VH, offering a more comprehensive representation of the scene.% The polarization in SAR images refers to the orientation of the plane in which the transmitted electromagnetic wave oscillates.

The data coming from satellite images is used by scientists for a wide range of applications including environmental protection~\cite{russwurm2023improved}, biodiversity estimation~\cite{kellenberger2019half} and demographic studies~\cite{rousse2024domain}. This data is also used by the public or journalists to identify events, conflicts, or the climate crisis~\cite{BurningSkies24}.
However, it is time-consuming to extract information from remote sensing images. This work is performed by experts and often involves manual work, which can be a limiting factor considering the growth of data volumes. In addition, the extraction of information from satellite imagery is often limited to optical sensors operating in the visible spectrum, as they are easier to interpret.
Indeed, interpreting SAR data is challenging due to geometric variability (with phenomena such as shortening, layout, and shadowing) and SAR image statistics called speckle.

To facilitate the extraction of information from remote sensing data, authors of~\cite{lobry2020rsvqa} have proposed a new task in which the objective is to provide an answer to an open-ended question, expressed in natural language, about remote sensing images. This task is known as Remote Sensing Visual Question Answering (RSVQA) and originates from Visual Question Answering (VQA)~\cite{antol2015vqa}. In~\cite{lobry2020rsvqa}, the authors provide two datasets and a method that separately extracts textual and visual features from the questions and images. These features are then combined and passed to a multi-layer perceptron to choose the most appropriate answer to the question among a set of pre-defined ones. %using pointwise multiplication 
%into a final vector used to make the prediction and answer the question.

Although this method shows promising results with optical images, it is not interpretable. In Prompt-RSVQA~\cite{chappuis2022prompt}, this issue is addressed by dividing the process into two phases. In the first phase, the model identifies relevant information in the images to answer the questions. In the second phase, features are extracted from the question and combined with the class names identified in the images using a language model. %The transformer's output is a classification vector, where the highest value corresponds to the most probable answer. 
This two-step method allows for a more detailed study of the classes detected in the images and a better understanding of potential prediction errors.

Both methods rely exclusively on optical images, much like the majority of the current state-of-the-art in RSVQA, with the exception of~\cite{tosato2024sarimproversvqaperformance}. 
This trend is also prevalent across other tasks. The reliance on optical images stems from the greater complexity of SAR images, which has limited their usage in deep-learning based methodologies~\cite{zhu2021deep}. %YOU WOULD NEED A CITATION FOR THAT Especially in the field of language, the research is still in its early stages.

However, a fusion of SAR with optical images can leverage the unique strengths of each type of imagery. SAR images provide complementary information, such as detailed texture and surface characteristics. In addition, the content of optical images may not be visible at all due to atmospheric conditions or poor lighting. Integrating SAR with optical images can enhance overall analysis and interpretation by offering a comprehensive view of the scene. This combined approach has been shown to improve various applications, including object detection and land classification ~\cite{joshi2016review}.

In this work, we explore the use of SAR imagery in two RSVQA pipelines, End-to-End RSVQA and Prompt-RSVQA. We also study the effects of different fusion methods to combine optical and SAR imagery for RSVQA. %Finally, an analysis is carried out to understand how SAR imagery and the various fusions are useful for our task.

The paper is structured as follows: Section~\ref{relatedw} reviews RSVQA, deep learning with SAR, and data fusion. Section~\ref{method} presents our pipelines and fusion methods, followed by our dataset in Section~\ref{data}. Section~\ref{ssec:vqa_metric} details evaluation metrics, with experiments and results in Section~\ref{experiments}. We discuss findings in Section~\ref{discussion} and conclude in Section~\ref{conclusion} with insights on SAR in RSVQA and future directions.

\section{Related Works}\label{relatedw}
RSVQA has been first introduced in~\cite{lobry2020rsvqa}. 
Since then, many methods have been proposed, including approaches for identifying the best language models~\cite {chappuis2022language}, fusion method between text and images~\cite{chappuis2021find}, using segmentation maps to guide the fusion~\cite{tosato2024segmentation}, and using text-based data augmentation~\cite{yuan2023multilingual}. 
The applications of RSVQA have broadened, with models now capable of providing diverse answers. For example,~\cite{chappuis2023multi} explores an object detection model based on textual questions, while~\cite{guo2024remote} introduces a model that answers questions requiring segmentation maps, vector maps, and object counting through a VQA interface, showcasing the versatility of RSVQA.

Research in RSVQA has predominantly focused on optical images, with SAR images being introduced only recently. The first study to incorporate SAR images~\cite{tosato2024sarimproversvqaperformance}, addressed land classification. Subsequent works explored questions related to scattering patterns and backscattering mechanisms~\cite{aghababaei2024visual}, boat detection~\cite{wang2024visual}, and comparisons between the results got using optical images, obscured by clouds or captured at night, and SAR's one~\cite{zhao2024text}.

\begin{comment}
Recent efforts in the remote sensing community have explored deep learning with SAR data for tasks such as object detection~\cite{chen2014sar}, despeckling~\cite{chierchia2017sar}, and image generation~\cite{letheule2023automatic}. 
While SAR and text interaction is still emerging due to the complexity of SAR data, only a few works have explored this area. These studies have found that SAR images are effective in describing object relationships, such as proximity and density (e.g., "few," "many"), but struggle with target size and object counting~\cite{zhao2022exploring}. Additionally, text is used with SAR to generate images, address dataset imbalances, and improve data sample diversity~\cite{wang2024improved}.
\end{comment}

Recent efforts in the remote sensing community have explored deep learning with SAR data. Hence, works on object detection~\cite{chen2014sar}, despeckling~\cite{chierchia2017sar}, volcano deformation detection~\cite{valade2019towards} and images generation~\cite{letheule2023automatic} have been proposed. In general, all techniques exclusively applied to optical imaging are now also extended to SAR imaging~\cite{zhu2021deep}. 

The interaction between SAR images and text is still in its early stages, primarily due to the inherent complexity of SAR data. While deep learning has shown limited effectiveness in tasks such as describing target sizes or counting objects~\cite{zhao2022exploring}, it performs better when identifying the relative positions of targets—such as indicating proximity to other objects—or providing density-based descriptions like "few," "many," or "a lot." Additionally, text has been used in conjunction with SAR data to generate synthetic images, helping to mitigate dataset imbalances and address the challenge of limited training samples~\cite{wang2024improved}.

One of the reasons for the predominant use of optical images is that they represent the visible spectrum. This makes them ideal for tasks such as object recognition~\cite{guo2020learning}, though they are limited by weather and lighting conditions. In contrast, SAR images, generated using radar signals, provide structural information and operate effectively in all weather and lighting conditions but are harder to interpret due to their complex patterns. Combining the optical and SAR modalities can leverage the strengths of both modalities, with optical images providing visual information and SAR images offering structural insights, resulting in a more comprehensive and accurate representation of the scene~\cite{hughes2020deep}.

Various fusion techniques have been explored to combine optical and SAR images in deep learning pipelines. Early fusion, also known as raw data fusion, merges optical and SAR images before feature extraction, and has been applied in pansharpening~\cite{yilmaz2022theoretical} and Digital Surface Model (DSM) generation from multiangular images~\cite{pacifici2008urban}. This approach is especially useful when the images are captured under similar conditions, such as using the same sensor or when the acquisition times are closely aligned. Halfway fusion, or feature-level fusion, extracts features from both modalities separately and then combines them for tasks like classification and change detection~\cite{lee2018pedestrian}, offering a richer set of characteristics. Late fusion, or decision-level fusion, operates at the highest semantic level by merging the outputs of single-modality pipelines and is particularly effective in classification tasks~\cite{licciardi2009decision}.
The choice between halfway and late fusion, however, remains task-dependent. In classification, it is challenging to determine which technique performs better, as studies report that halfway fusion sometimes yields superior results~\cite{gadzicki2020early,liu2016multispectral}, while late fusion excels in other cases~\cite{wagner2016multispectral}. As noted in~\cite{dalla2015challenges}, the performance of either method is closely tied to the specifics of the problem and dataset. %In~\cite{hu2024opt}, the authors demonstrate that a combination of all three fusion types can be advantageous, depending on the application.

Although some recent works have explored RSVQA with SAR images, all focus on very high-resolution SAR images~\cite{aghababaei2024visual}~\cite{wang2024visual}~\cite{zhao2024text}. 
%This makes it difficult to compare the results with those obtained from optical images captured in clear weather and daylight conditions. 
This makes it difficult to compare the results with those obtained from optical images captured in clear weather and daylight conditions, especially when the questions differ for SAR and optical images and the images do not represent the same areas~\cite{zhang2024earthgpt}.

While it is known that SAR images perform better at night and on cloudy days, there has been no study in the RSVQA field on whether they provide unique information that optical images do not. Our work aims to address this gap by building on the findings of~\cite{tosato2024sarimproversvqaperformance}, examining the performance of two RSVQA pipelines when using SAR images alone or in combination with optical data.

\begin{comment}
It appears that there is a gap in the state-of-the-art regarding the use of SAR images for RSVQA, even though SAR has been demonstrated to be useful in several fields due to the complementary information it provides compared to optical images~\cite{hughes2020deep}. We propose to extend the work done in~\cite{tosato2024sarimproversvqaperformance} by studying the behaviour of two pipelines when using SAR images alone or in addition to optical data. 
\end{comment}

\section{Method} \label{method}
\begin{figure*}
    \centering
    \includegraphics[width=\textwidth]{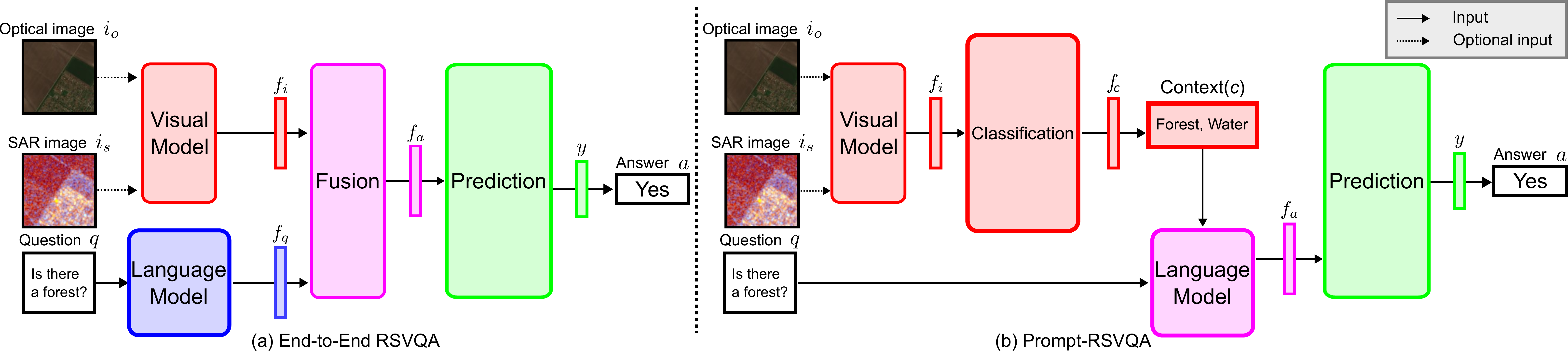}
    \caption{\label{fig:methods}
    The two pipelines proposed in this work. In both pipelines, we extract the visual information (an abstract visual feature vector $f_i$ for End-to-End RSVQA or a classification vector $f_c$ for Prompt-RSVQA) from one (i.e. $i = i_o$ or $i = i_s$) or two modalities (i.e. $i = (i_o, i_s)$). The End-to-End RSVQA pipeline processes both modalities separately. In this pipeline, we extract a feature vector $f_q$ from the question $q$. Both feature vectors are merged in a multi-modal vector $f_a$. In the Prompt-RSVQA pipeline, we convert the multi-label classification vector $f_c$ to text through a thresholding operation. This text is passed along the question to obtain a feature vector $f_a$ representing both the images and the question. In both pipelines, the multi-modal feature vector $f_a$ is used to predict the most likely answer $a$ from a set of possible answers $\mathcal{A}$.}
    \label{fig:enter-label}
\end{figure*}

\begin{comment}
This section presents the method carried out in this work. We aim to study the integration of SAR images into a neural network for RSVQA, to establish its usability for this task. 
To achieve this, we analyse the robustness of two pipelines. Once the best method has been defined, we study which method of merging the information present in the various modes is the most suitable for our task. 
\end{comment}
%Put more focus on VQA
% Problem definition: Our objective is to obtain answer y from ... (put in front that you have SAR and opt images as input)
The RSVQA task aims at providing an answer $a$ to a question $q$ from an image $i$. As seen in Section~\ref{relatedw}, recent works on RSVQA only focused on the case in which $i = i_o$ is an optical image (with multiple spectral channels $n_o$). In this work, we introduce a SAR image $i_s$ (with multiple polarization channels $n_s$) to the problem setting. Hence, we formulate our task as providing an answer $a$ to a question $q$ from either a single modality (i.e. $i = i_s$ or $i = i_o$) or a pair of optical and SAR images $i = (i_o, i_s)$. Similarly to other works, we frame the RSVQA task as a classification problem, where the answer $a$ is predicted among a set of pre-defined answers $\mathcal{A}$.
%In RSVQA, the aim is to feed a neural network with an optical image $i_o$ and a question $q$ into a model and have as output a correct answer $a$. 
%Our method aims to introduce a SAR image $i_s$ in this pipeline in the most effective way.

%IMPORTANT: Do not talk about results yet. So you can not say yet "after identifying which method is best".
To predict the right answer, we propose two methods shown in Figure~\ref{fig:methods}: the first one is an End-to-End method (\textbf{End-to-End RSVQA}, described in Sub-section~\ref{ssec:eteRSVQA}). In this method, abstract features are extracted from the images $i$, as well as from the question $q$. These features are then merged to predict the answer $a$. The second one (\textbf{Prompt-RSVQA}, described in Sub-section~\ref{ssec:prompt}) is inspired by~\cite{chappuis2022prompt}. In this method, we first extract semantic information from the images. This is done through a separately trained multi-label classification network. This semantic information is then passed, along with the question $q$ to a language model to predict $a$. In this work, we also experiment with different methods for the extraction of the visual feature vector, with the proposition of three fusion mechanisms. The visual feature extractors are presented in Sub-section~\ref{ssec:fusion}.

\subsection{End-to-End RSVQA}
\label{ssec:eteRSVQA}
In the End-to-End RSVQA pipeline, both the images $i$ and the question $q$ are processed separately. 
The different encoders used to obtain $f_i$, the feature vector of the images $i$, are described in section~\ref{ssec:fusion}. To obtain $f_q$, the feature vector of the question, we use a Recurrent Neural Network (RNN). We add a fully connected layer to map both vectors to new vectors $f'_i$ and $f'_q$ both of dimension $n_a$. Both of these vectors are merged into a new vector, $f_a$ through a point-wise multiplication. While the point-wise multiplication is a simple and fixed operation, the fully connected layers applied on $f_i$ and $f_q$ leave a degree of freedom for the network to reorganize the information in a way that helps the fusion of both vectors. Finally, the vector $f_a$, representing both the visual and textual information, is used as an input to a multi-layer perceptron to predict the final answer $a$. This MLP outputs a vector $y$, of size $|\mathcal{A}|$, giving a score for each possible answer from $\mathcal{A}$.
%In the End-to-End RSVQA, the image $i$ is processed through a CNN where the last average pooling layer and fully connected layer are replaced by a per-pixel perceptron, which produces the vector $vi_{1}$.
%Simultaneously, the question $q$ is processed by an RNN, yielding a vector $vq_{1}$. $vi_{1}$ and $vq_{1}$ are transformed, through a linear layer, into two new vectors $vi_{2}$ and  $vq_{2}$ of the same dimension. 
%The vectors $vi_{2}$ and $vq_{2}$, now of the same size, are fused through pointwise multiplication to create the vector $vt_{1}$. This vector is then fed into a neural network, producing a final output layer $\mathcal{A}$ representing the number of possible answers. In this manner, the prediction of the answer $a$ is treated as a classification problem.

% Introduce "prompt" (cite the paper of Christel, "we take inspiration from), introduce the different fusion methods in details (use the figure here)
\subsection{Prompt-RSVQA}
\label{ssec:prompt}
Prompt-RSVQA is organized into two stages. The first stage aims at extracting relevant semantic information from the image $i$. After obtaining $f_i$ (see Sub-section~\ref{ssec:fusion}), we predict $f_c$ a vector that represents classification scores %In this work, the set of classes is obtained 
through a multi-layer perceptron. More specifically, after applying a threshold on the prediction vector $f_c$, we obtain the set of classes describing the content of the image, which are concatenated in a text-based list called context $c$. Because this operation is non-differentiable, this first part of the pipeline is trained separately.

The second stage of Prompt-RSVQA takes as an input the question $q$ and the context $c$ extracted from $i$ in the first phase. Both of these texts are concatenated and fed into a transformer-based language model. To keep a similar setting to the pipeline of End-to-End RSVQA, we only use encoder layers and extract the feature vector $f_a$. This vector represents the information of the question and the context extracted from the image. Similarly, we use a multi-layer perceptron to predict the vector $y$ from $f_a$.

\begin{figure*}
    \centering
    \includegraphics[width=\textwidth]{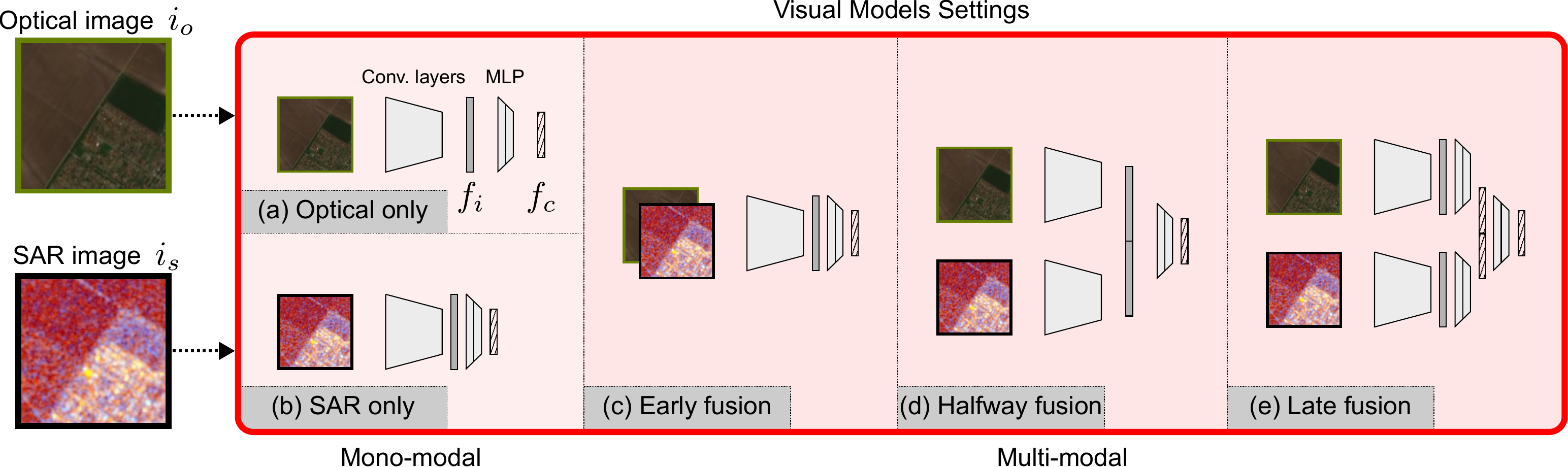}
    \caption{\label{fig:visual_models}The different visual models studied in this work. (a) and (b) are mono-modalities models. In both cases, we use a CNN to extract abstract features $f_i$ (which can be used in End-to-End RSVQA) or high-level information $f_c$ (in our case, classes obtained after a MLP applied on the output of the convolutional layers, which can be used in Prompt-RSVQA). Similarly, the early fusion mechanism (c) uses a single CNN. However, in this setting, both $i_o$ and $i_s$ are stacked at the input of the CNN. In the halfway fusion (d), we concatenate the output of the convolutional layers before doing the classification. Finally, the late fusion mechanism adds a MLP to the classification scores obtained from each modality separately.}
\end{figure*}

\subsection{Visual models}\label{ssec:fusion}
In this work, we examine different strategies to obtain a representation of the information (either abstract, $f_i$ for End-to-End RSVQA, or a classification $f_c$ in the case of Prompt-RSVQA) contained in the images $(i_o, i_s)$. To this end, we propose to examine five methods presented in Figure~\ref{fig:visual_models}: optical only, SAR only, early fusion, halfway fusion and late fusion.
In the mono-modal strategies (i.e. optical only or SAR only), we use a CNN as one would for classification. First, a series of convolutional layers are applied to obtain the visual feature vector $f_i$. This vector is then passed to a MLP which outputs a prediction vector $f_c$ indicating the scores for each possible class.

%Once we have studied the robustness of the two methods and assessed which one gives better results with SAR, we use this method to test which fusion mechanism is best within the RSVQA context.\\
We first propose an \textbf{early fusion} method  (Figure~\ref{fig:visual_models}(c)). In this approach, the two modalities $i_o$ and $i_s$, with $n_o$ and $n_s$ channels respectively are stacked to create a single image with $n_o+n_s$ channels, which is then fed into the model. Similarly to the mono-modal strategies, we then use a CNN to extract the visual feature vector $f_i$ and a MLP to obtain the classification scores $f_c$.

The second fusion method, \textbf{halfway fusion} (Figure~\ref{fig:visual_models}(d)), processes the two modalities separately through different CNNs. Before the final step, their feature maps are concatenated into a single feature map $f_i$, then passed to an MLP to obtain $f_c$.

The final method is named \textbf{late fusion} and is presented in Figure~\ref{fig:visual_models}(e). With this strategy, the two modalities are processed separately through different CNNs. 
A vector representing classification scores is produced for each modality. These two vectors are then concatenated and passed to a MLP which finally produces the final decision vector, $f_c$.

\section{Data} \label{data}
\begin{figure*}
    \centering
    \includegraphics[width=\textwidth]{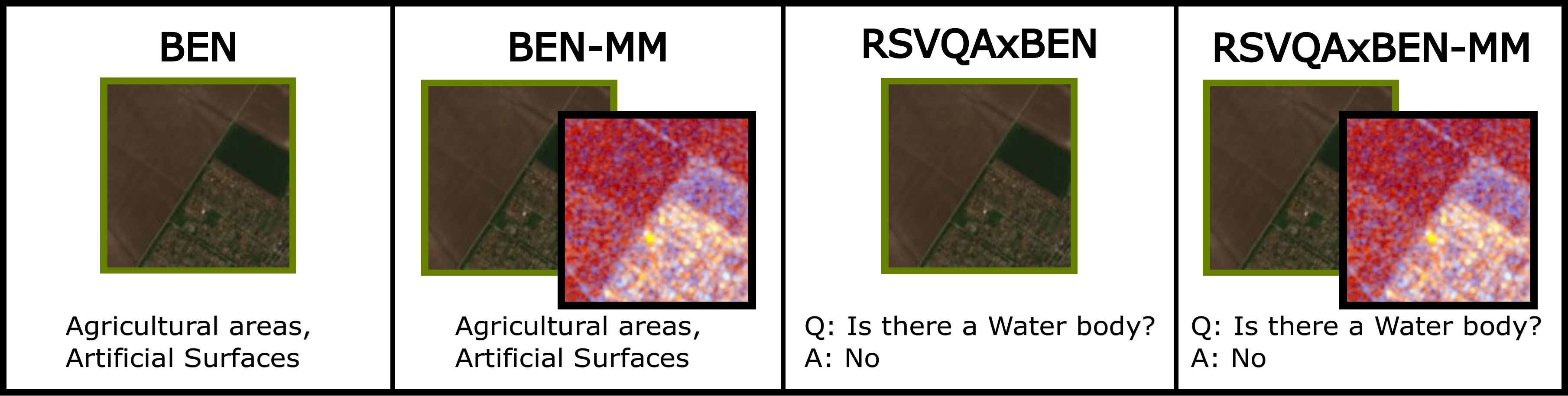}
    \caption{Workflow from the initial BEN dataset containing the Sentinel-2 optical images and their CLC classes, to BEN-MM-61 where the respective Sentinel-1 images were added. We then move on to RSVQAxBEN in which questions and answers related to the optical images have been proposed. Finally, the dataset we propose in which such questions and answers are linked to both optical and SAR images.}
    \label{fig:datasets}
\end{figure*}

This section introduces a new dataset for multi-modal RSVQA: \textbf{RSVQAxBEN-MM}. This dataset is derived from three other datasets: BEN and BEN-MM, discussed in Sub-section~\ref{ssec:BEN} and RSVQAxBEN, discussed in Sub-section~\ref{ssec:RSVQAxBEN}. A sample of each of these datasets is shown in Figure~\ref{fig:datasets}.

\subsection{Land-cover datasets: BEN and BEN-MM}\label{ssec:BEN}
BigEarthNet~\cite{sumbul2019bigearthnet} (BEN) dataset is composed of 590'326 patches of Sentinel-2 images with 12 channels, acquired from over 10 European countries. Each patch is matched with the classes corresponding to the CORINE Land Cover (CLC) map of 2018.
In the CLC nomenclature, the land cover classes are presented on three increasingly specific hierarchical levels. Level L1 introduces 5 more generic classes (e.g. Agricultural areas, Water bodies). Level L2 introduces 15 sub-classes (e.g. Arable land, Inland water). Finally, level  L3 presents 44 classes at the finest level (e.g. Rice fields, Water courses). Over the three levels of information, the CLC nomenclature introduces 64 classes. In BEN, some classes have been deleted and others have been fused reaching a total of 19 classes. The train (60\% of the patches), validation (20\% of the patches) and test sets (20\% of the patches) are defined randomly. 

An extension of BEN, called BigEarthNet-Multi Modality dataset~\cite{sumbul2021bigearthnet} (BEN-MM), has been later released adding to each Sentinel-2 patch of the original dataset the corresponding Sentinel-1 image, in the two polarisation channels (VV and VH) in dB. %Note that this extension does not include new classes, and keeps the same 19 classes nomenclature as BEN. Moreover, the same train, validation and test sets distribution is kept.
In this work, we also use another version of the BEN-MM dataset, with 61 classes as labels (see Section~\ref{ssec:RSVQAxBEN}). In the rest of this manuscript, we refer to this dataset as BEN-MM-61.

\subsection{RSVQA dataset: RSVQAxBEN}\label{ssec:RSVQAxBEN}
The \textbf{RSVQAxBEN dataset} is derived from BEN. Each of the 590,326 Sentinel-2 RGB patches is paired with 25 question/answer pairs, totaling 14,758,150 image/question/answer triplets.

The questions are based on each image's CLC labels. As opposed to BEN, the 64 original classes of the CLC nomenclature are kept. However, two classes with the same name at different hierarchy levels (\textit{water bodies} and \textit{pastures}) are counted as a single label and the category \textit{Glaciers and perpetual snow} is removed, leading to a total of 61 classes.

Questions fall into two categories: \textit{yes/no questions}, in which the answer is 'yes' or 'no' (e.g. 'Is there a forest or water in this image?') and \textit{land cover questions} in which the answer is a list of classes (e.g. 'Besides forest, what classes are present in the image?'). The dataset is heavily unbalanced, with  80.7\% of the questions being a \textit{yes/no question}. In \textit{yes/no questions}, conjunctions such as 'and' and 'or', adding a difficulty level to the question, are present in 72.3\% of the cases. In 27.1\% of the questions, two conjunctions are found. As opposed to BEN, the dataset split is divided according to the spatial location of the image: The westernmost patches are placed in the training set (66\% of the dataset); the easternmost patches are put in the test set (23\% of the dataset); finally, the other patches, representing 11\% of the dataset, are put in the validation set. This choice allows the models to avoid biases due to geographical location. This split minimizes location-based biases but introduces a geographical generalization challenge, as some classes vary visually between Western and Eastern Europe. %However, it implies a difficult geographical generalization problem, as some classes present different visual representations in Western Europe (i.e. the train set) and in Eastern Europe (i.e. the test set).

\subsection{Proposed dataset: RSVQAxBEN-MM}
This work introduces the RSVQAxBEN-MM dataset. This dataset adds the SAR images from BEN-MM to the question/answer of RSVQAxBEN. An example is shown in Figure~\ref{fig:datasets}(d). Because it is derived from RSVQAxBEN, this dataset keeps the same choices, in terms of classes of interest and train / validation / test sets distribution.
Each of the images / question / answer triplets is composed of the Sentinel-2 RGB patch as in RSVQAxBEN and adds the corresponding Sentinel-1 patch. The provided Sentinel-1 patches are composed of the VV, and VH channels in dB and the ratio between the two, providing $n_s=3$ SAR channels.

We use three channels, as combining VV and VH polarizations improves the separability of vegetation and artificial structures~\cite{lee2001quantitative}, while the VV/VH ratio effectively distinguishes volume from surface scattering~\cite{tosato2024sarimproversvqaperformance,park2008integration}. Both enhancing land cover classification in RSVQA tasks. In this context, with images expressed in dB scale, the ratio channel is calculated as the difference between VV and VH, normalized between 0 and 1.  %\textcolor{orange}{(Reviewer 2, comment 2/3)}

The RGB representation of the SAR image is obtained by assigning VV to the R channel, VH to the green channel and the ratio to the blue channel. %\textcolor{orange}{(Reviewer 2, comment 4)}

\subsection{Dataset evaluation)
} 
\label{limitations}
While RSVQAxBEN-MM is the first dataset proposing SAR and optical modalities for RSVQA, it presents some limitations discussed in this section.
The first limitation concerns the distribution of the CLC classes. Across the different patches, the 61 classes are strongly imbalanced. 
Among the three levels, the six most represented classes (Agricultural areas and Forest and Seminatural areas at L1, Arable land, Heterogeneous agricultural areas and Forests at L2, and Non-irrigated arable land at L3) together account for 54\% of the 4'457'279 class occurrences in BEN-MM-61. This imbalance makes the classification task harder. In particular, it has been shown that since the questions and answers are based on the classes in each image, the use of weighted-losses during the training brings no benefit in VQA~\cite{tosato2024sarimproversvqaperformance}.

The second limitation concerns the biases present in the answers.  To evaluate the biases in the dataset some scores are introduced in~\cite{chappuis2023curse}, namely: Uniform distribution, Prior distribution, and  $L_{B_{score}}$.
The uniform distribution is calculated as the inverse of the number of unique answers $A_{\text{unique}}$:

\begin{equation}
    \text{Uniform} = \frac{1}{A_{\text{unique}}} \quad.
\end{equation}
The Prior distribution is calculated as the ratio between the number of samples with the most common answer $A_{\text{common}}$ and the total number of samples ($N$):
\begin{equation}
    \text{Prior} = \frac{A_{\text{common}}}{N} \quad.
\end{equation}
Finally, assuming that \text{Uniform} does not equal 1 in a realistic scenario, the $L_{B_{\text{score}}}$ can be calculated:
\begin{equation}
    L_{B_{\text{score}}} = \frac{\text{Prior} - \text{Uniform}}{1 - \text{Uniform}} \quad.
\end{equation}
All these scores work in a range of values between 0 and 1.
Ideally, a perfect RSVQA dataset would have ${A_{unique}}=\dfrac{N}{A_{common}}$, which means that each answer has the same number of occurrences. In this perfect scenario, the $L_{B_{score}}$ would be equal to 0. In Table~\ref{rsvqabiasscores} the scores are applied to RSVQAxBEN-MM. When the question type is \textit{All}, it measures the bias by treating the entire dataset as a single group, calculating how much the model favours the most common answer across all questions combined.
These scores are also computed by question category, which are then combined on the whole dataset. %\textit{Average} combination, the bias scores are averaged, giving equal weight to each question category regardless of number of the questions.
%The \textit{Overall} combinaison aggregates bias across categories by weighting the bias with weights proportional to the number of questions in each category, providing a more comprehensive view of the model's bias.
 
We can observe in \textit{All} and \textit{Land Cover} in Table~\ref{rsvqabiasscores}, that the Uniform distribution has a very low value, which means that there are many different types of answers. Instead in \textit{Yes/No} the Uniform distribution has a value of 0.50 since there are only two types of answers. 

Prior in \textit{All} has a value of 0.52, which means that more than half of the answers are ‘no’ and is a direct result of the fact that in \textit{Yes/No} as many as 63\% of the answers are ‘no’. 
The dataset appears to be biased towards the most common answers, particularly when the dataset is viewed as a whole.
%However, the \textit{Overall} score shows that while there is still some bias, it’s less severe when considering the dataset structure and the distribution of samples across categories.

\begin{table*}
\centering
\small
\begin{tabular}{|l|c|c|c|c|c|c|}
\hline
\textbf{Question Type} & \textbf{\#samples} & \textbf{\#answers} & \textbf{Most common} & \textbf{Prior} & \textbf{Uniform} & $\mathbf{L_{B_{score}}}$\\
\hline

\textbf{All} & 2'953'125 & 26'875 & no & 0.52 & 0.00004 & 0.52 \\
 \textbf{Land cover} & 529'413 & 26'873 & None & 0.13 & 0.00004 & 0.13 \\
 \textbf{Yes/No} & 2'423'712 & 2 & no & 0.63 & 0.50 & 0.26 \\
 \hline
\end{tabular}
\caption{Analysis of RSVQAxBEN test set done in~\cite{chappuis2023curse} using the Prior, Uniform and $L_{B_{score}}$ scores.}
\label{rsvqabiasscores}
\end{table*}
%Despite these limitations, it must be mentioned that in reality, the classes present are not balanced, let alone the words used to answer the questions.
Despite these limitations, it is important to note that the distribution of classes and therefore answers is naturally unbalanced in reality as well.

\section{Performance evaluation}\label{ssec:vqa_metric}
To evaluate our work, we use a VQA metric described in~\autoref{ssec:metric:VQA}. In addition, we introduce metrics for classification in~\autoref{ssec:metric:classif} and~\autoref{ssec:metric:classif_vqa} that we use for the evaluation in the context prediction for Prompt-RSVQA.

\subsection{Classification}
\label{ssec:metric:classif}
\textbf{F1 Score:} is defined as the harmonic mean of the precision (P) and recall (R), where an F1 score reaches its best value at 1 and worst score at 0: 
\begin{equation}
\textrm{F1} = 2 \cdot \frac{P \cdot R}{P+R},
\end{equation} \label{eq1}
with
\begin{equation}
    P = \frac{TP}{TP + FP}\quad R = \frac{TP}{TP + FN},
\end{equation}
and $TP$ the number of true positive predictions, $FP$ the number of false positives and $FN$ the number of false negatives.

The F1 score is computed for each class. In the following, we report the F1-micro score and the F1-average score.
The first one counts the total true positives, false negatives and false positives.
While the latter is computed as a weighted arithmetic mean (with a per-class weight corresponding to the number of positives of the class) of the per-class F1.

\subsection{From classification to VQA}
\label{ssec:metric:classif_vqa}
\textbf{Match Ratio (MR):} 
The MR computes the fraction of correctly classified samples, i.e. the samples whose predicted labels exactly correspond to the ground truth labels. This gives the following for \textit{Q} samples:
\begin{equation}
    MR = \frac{1}{Q}\sum_{i=1}^Q I(f_{c_{i}}=\hat{f}_{c_{i}})\,,\label{eq3}
\end{equation}
where $f_{c_{i}}$ represents the actual labels (as a one-hot vector) for the $i_{th}$ sample, while $\hat{f}_{c_{i}}$ represents the labels predicted by the model for the same sample after thresholding. The identity function $I$ is 1 for an exact match and 0 otherwise.

\textbf{Hamming Distance (HD): }
The HD is defined as the number of classes with a different prediction than the ground truth. It is defined for \textit{Q} samples and \textit{N} land cover categories as:
\begin{equation}
    HD = \frac{1}{Q}\sum_{i=1}^Q \sum_{j=1}^N I(f_{c_{{ij}}}=\hat{f}_{c_{ij}})\label{eq4}\,,
\end{equation}
where $f_{c_{{ij}}}$ and $\hat{f}_{c_{ij}}$ represent the prediction and ground truth of the $j^{th}$ land cover class for the $i^{th}$ sample.

\subsection{VQA}
\label{ssec:metric:VQA}
To evaluate VQA results, we define the percentage of correct answers as the accuracy. The global accuracy and per type of question (“Yes/No” or “Land cover” subsets) accuracy are provided. 

\section{Experiments}\label{experiments}
We evaluate our proposed method for RSVQA using optical images (Sentinel-2 RGB) and SAR images (VV and VH polarizations and their ratio). We first train image encoders on the BEN-MM-61 classification task and present the results in~\autoref{expclass}. We compare this encoder with an ImageNet pre-trained encoder for the End-to-End RSVQA pipeline in~\autoref{ssec:exp:e2e}. Finally, we present our results obtained with the Prompt-RSVQA pipeline in~\autoref{ssec:exp:prompt}.

\subsection{Classification}\label{expclass}
%The classification experiments on BEN-MM are a pre-requisit for the PROMPT-RSVQA methods, but training networks on the classification task using the BEN-MM dataset also enable to adapt the visual model part of the End-to-End RSVQA method to the dataset.
In the first classification experiments, optical and SAR images are used separately.
Two networks of different depths, ResNet-50 and ResNet-152 are compared to assess whether greater depth yields better results.
The last layer of the network is replaced with one that has the correct output number, i.e. 61. The sigmoid is used as an activation function. 
The performances of the single-modality encoders are presented in Table~\ref{tab:ClassBENMM} (rows a,b,c,d). 
The performances are evaluated on the classification task using F1-micro and F1-average score. In addition, we assess the performances of the encoders for the RSVQA task using MR and HD.

Based on the single-modality results, we compare the three fusion methods introduced in~\autoref{ssec:fusion} (early fusion, halfway fusion and late fusion) with ResNet-50. The results are presented in the row e,f,g of Table ~\ref{tab:ClassBENMM}.

In the early fusion, optical and SAR images are concatenated and inserted into a ResNet-50 pre-trained on ImageNet. The first layer of the ResNet is modified to have 6 channels as input instead of 3. 
In this layer, the weights are initialised using the weights resulting from the optical-only training of the ResNet-50 on BEN-MM-61 for the optical channels and the weights resulting from the SAR-only training of the ResNet-50 on BEN-MM-61 for the SAR channel. %Each epoch takes about three hours.

In the halfway fusion, each modality is inserted in a frozen ResNet-50 pre-trained on the BEN-MM-61 classification task, with parameters set as in Table~\ref{tab:ClassBENMM}[b,d]. The last layer of each network is deleted and the feature maps of the two modalities are concatenated. An average pooling, flattening, linear layer and sigmoid are applied to the concatenated feature maps to finally have the prediction of the classes. This last linear layer is retrained on the classification task with BEN-MM-61. %Each epoch takes about four hours.

The late fusion takes as input the two optical and SAR images and inserts them into two different frozen ResNet-50 fully pre-trained on the classification task with the parameters set as in~\ref{tab:ClassBENMM}[b,d].
The two decision vectors, before thresholding, are concatenated and fed into an MLP, which generates the final vector representing the detected classes in the images. Considering \( n_c \) as the number of classes, the two vectors are concatenated after applying the sigmoid function, resulting in an input of size \( n_c \times 2 \) for the MLP. The MLP consists of three layers that progressively transform the vector size.
The final output, has a dimention of 61, representing the probabilities of the predicted classes based on the combined information from both modalities. This MLP is also retrained on a classification task using BEN-MM-61.
%Since the models for SAR and optical images are trained separately, each epoch takes approximately one and a half hours. The additional MLP requires 10 minutes per epoch to train.
 
We train all the models using Adam as an optimizer, a learning rate of $10^{-6}$, a batch size of 64 and a binary cross-entropy as the loss. The number of epochs used to train each model is displayed in Table~\ref{tab:ClassBENMM}. In this work, all of the models are trained with an Nvidia V100 with 16GB GPU. The number of trainable hyperparameters per method is present in Table~\ref{param}.
\begin{table}[h!]
\begin{center}
{\small
\begin{tabular}{|l|r|}
\hline
Model & Trainable Parameters \\ \hline
S2/S1 ResNet-152 & 58'268'797 \\
S2/S1 ResNet-50  & 23'633'021 \\
Early Fusion     & 23'642'429 \\
Halfway Fusion   & 249'917    \\
Late Fusion      & 67'405     \\ \hline
\end{tabular}
}
\end{center}
\caption{Number of trainable parameters per model.
}
\label{param}
\end{table}
\subsection{End-to-End RSVQA}\label{ssec:exp:e2e}
In~\cite{lobry2020rsvqa}, the End-to-End RSVQA was tested only with optical data. In this study, we assess the robustness of the method using SAR images as inputs, and compare it to the results of optical only.
For each modality, we compare ResNet-152 and ResNet-50 to extract features compatible with text. In both cases, an MLP composed of two layers is added at the end, after the application of a Hyperbolic Tangent function and a dropout of 0.5. 
First, the networks are frozen and their weights are initialized either with the pre-training on ImageNet or the classification task on BEN-MM-61 as described in section~\ref{expclass}. 
In addition, for ResNet-50 initialized on ImageNet only, the network is unfrozen during the full End-to-End pipeline training. The results of the 10 experiments are presented in Table~\ref{tab:RSVQAexp}. 
%{In Table~\ref{tab:RSVQAexp} the different settings for the experiments are presented.We start using a ResNet-152 pre-trained on ImageNet and in a classification task on the same dataset BEN-MM. Then we try decreasing the capacity of the network using ResNet-50 pre-trained on ImageNet. After we use a ResNet-50 pre-trained on a classification task, eventually we train it from scratch during the End-to-End pipeline. Finally, we try all the described experiments using SAR images instead of optical, for a total of 10 experiments on the End-to-End RSVQA pipeline.}

We trained all these models using Adam as an optimizer, a learning rate of $10^{-6}$, a batch size of 1024 and a cross-entropy as the loss for 20 epochs.

\subsection{Prompt-RSVQA}\label{ssec:exp:prompt}
In Prompt-RSVQA, the visual model extracts classes that are transformed in text to help the language model answer the question. Thus, the performance of the Prompt-RSVQA pipeline depends strongly on that of the visual model. DistilBERT is used as the LLM. After applying the Sigmoid activation function and a dropout of 0.5, two layers are added at the end to match the number of possible answers.
The seven experiments conducted on the classification task are compared in the Prompt-RSVQA pipeline. Their settings are summarized in Table ~\ref{tab:prompt}.

\section{Results and discussion} \label{discussion}
\subsection{Classification}\label{classificationdiscussion}
By comparing the classification results for optical and SAR images in Table~\ref{tab:ClassBENMM}[a,b] and Table~\ref{tab:ClassBENMM}[c,d], we can see that the F1-micro score is 6 to 7\% higher using optical images than using SAR images. This highlights the smaller discrimination capacity of SAR images, especially for classes such as green urban areas and wetlands, and in particular inland wetlands.

Table~\ref{tab:ClassBENMM}[a,b,c,d] also shows that increasing the depth of the neural network, while keeping the same hyper-parameters and the same training time, does not lead to an improvement in classification. The performance of the ResNet-152 network is below the one of ResNet-50 for both modalities. 
In Table~\ref{tab:ClassBENMM}[a,b] we can verify that optical results lose  1.6\% % of F1-micro score when using ResNet-152 instead of ResNet-50. SAR results also decrease but only by 0.03\%.
Indeed, since our classification problem does not require such a complex representation, the additional depth of ResNet-152 may not provide significant benefits. On the contrary, it could worsen the results~\cite{saha2021space}. Moreover, the greater depth of ResNet-152 could limit its generalization capacities, decreasing its performance since the training set and our test set are geographically separated. This generalization issue could be greater in the case of optical data. 
Our results are in line with those obtained in~\cite{sumbul2019bigearthnet}, in which the shallowest network obtains the best results. 

\begin{comment}
{
We think this is due to two main reasons: 
\begin{itemize}
    \item ResNet-50 is already a very deep and powerful network. So it can be that the additional depth of ResNet-152 may not provide significant benefits if the problem does not require such a complex representation. On the contrary, it could worsen the results~\cite{saha2021space}.
    \item Adding more layers to a neural network can lead to improvements in performance, but at some point, the gains may become marginal and the time in execution increases more and more~\cite{khan2018evaluating}.
\end{itemize}
The larger drop in accuracy for optical images when switching from ResNet-50 to ResNet-152 could be due to overfitting on the richer visual details of optical images, while SAR images, with their different data structure, are less affected. ResNet-152 may also struggle with generalization on optical data due to its greater depth, and the use of ImageNet-pretrained weights likely amplifies this issue for optical images.}
\end{comment}

\begin{table*}[ht]
    \centering
    \small
    \begin{tabular}{|c|>{\centering\arraybackslash}p{1.6cm}|>{\centering\arraybackslash}p{1.6cm}|c|>{\centering\arraybackslash}p{1.6cm}||
>{\centering\arraybackslash}p{1.6cm}|>{\centering\arraybackslash}p{1.6cm}|>{\centering\arraybackslash}p{1.6cm}|>{\centering\arraybackslash}p{1.6cm}|}
        \hline
        \textbf{} & 
        \textbf{Modality} & \textbf{Network} & \textbf{Pre-trained on} & \textbf{Epochs} & \textbf{HD} & \textbf{MR} & \textbf{F1-micro} & \textbf{F1-avg} \\
        \hline
         a&OPT & ResNet-152 & ImageNet & 24 & 4.06 & 12.7\% & 73.00\% & 78.21\%\\
        \hline
         b&OPT & ResNet-50 & ImageNet & 40 & 3.40 &  \textbf{15.6\%} & 74.60\% & 80.92\% \\
        \hline
        \hline
         c&SAR & ResNet-152 & ImageNet & 7 & 4.16	& 11.94\%& 67.23\% & 76.36\% \\
        \hline
         d&SAR & ResNet-50 & ImageNet & 11 & 4.24 & 11.9\% & 67.26\% & 77.13\% \\
        \hline
        \hline
         e&Early & ResNet-50 & BEN-MM/ImageNet* & 31 & 4.03 & 13.7\% & 73.89\% & 80.54\%\\
         & Fusion & & & & & & & \\
        \hline
         f&Halfway & ResNet-50 & BEN-MM & 12 & 3.54 & 13.44\% & 74.10\% & 80.82\%\\
         & Fusion & & & & & & & \\
        \hline
         g&Late & ResNet-50 & BEN-MM & 2 &  \textbf{3.35} & 14.3\% 
         & \textbf{75.00\%}& \textbf{81.21\%} \\
         & Fusion & & & & & & & \\
        \hline
    \end{tabular}
    \caption{
    Classification results on BEN-MM using ResNet-50/152, using a single modality as input (rows a-d) and both modalities (rows e-g).\\ * The network is pretrained on ImageNet, where the first layer is initialized using weights from BEN-MM models (rows b–d). See Section~\ref{expclass} for details.
    }
    \label{tab:ClassBENMM}
\end{table*}
Then, we study the best type of fusion for our task considering only ResNet-50, as it performs the best for each modality individually.
From Table~\ref{tab:ClassBENMM} we can see that in terms of F1-micro, early fusion has the worst fusion results, followed by halfway fusion. Both have also slightly worse results than the optical image alone. Only late fusion does better than SAR and optical alone, performing the best. 

The low results of early fusion are not surprising since it has been shown in~\cite{tosato2024sarimproversvqaperformance} that inserting both SAR and optical images in one model may lead to primarily relying on optical channels.
The fact that late fusion outperforms halfway fusion is not expected, since we have seen in Section~\ref{relatedw} that the relative performances of each fusion depend on the nature of the task, its complexity, as well as on the dataset.
Halfway fusion combines features extracted from each domain (optical and SAR) before classification. We believe that for our task, the features extracted from the two domains are not well aligned or containing redundant or conflicting information. The fusion may introduce noise rather than useful information.
On the contrary, late fusion combines the predictions of the two modalities rather than their features. This may simplify the integration process since it operates on higher-level abstractions (e.g., class probabilities) rather than complex and potentially incompatible feature spaces. Interestingly, despite this simpler approach and having the fewest trainable parameters, as shown in Table~\ref{param}, the late fusion strategy still yields the best results among the three fusion schemes.
Moreover, combining final predictions might reduce the impact of modality-specific noise. For instance, if one modality is particularly noisy or less informative for certain classes, its influence can be minimized when combined with the other modality’s more accurate predictions.
Another advantage of the late fusion mechanism comes from the fact that both classifiers can be designed and trained separately, allowing to take advantage of specific strategies for each modality.

The BEN-MM-61 dataset is heavily unbalanced, with 6 classes that represent 54\% of the overall dataset, namely: Agricultural areas, Forest and seminatural areas, Arable land, Heterogeneous agricultural areas, Forests, Non-irrigated arable land. 
Late fusion improves four of them compared to a model considering optical data only: \textit{Forest and seminatural areas, Arable land, Heterogeneous agricultural areas} and \textit{Non-irrigated arable land}.
Halfway fusion improves on \textit{Forest and seminatural areas, Forests} and \textit{Non-irrigated arable land}. Finally, Early fusion shows an improvement on \textit{Agricultural areas} only.

\begin{figure*}[t!]
    \centering
    \includegraphics[width = \textwidth]{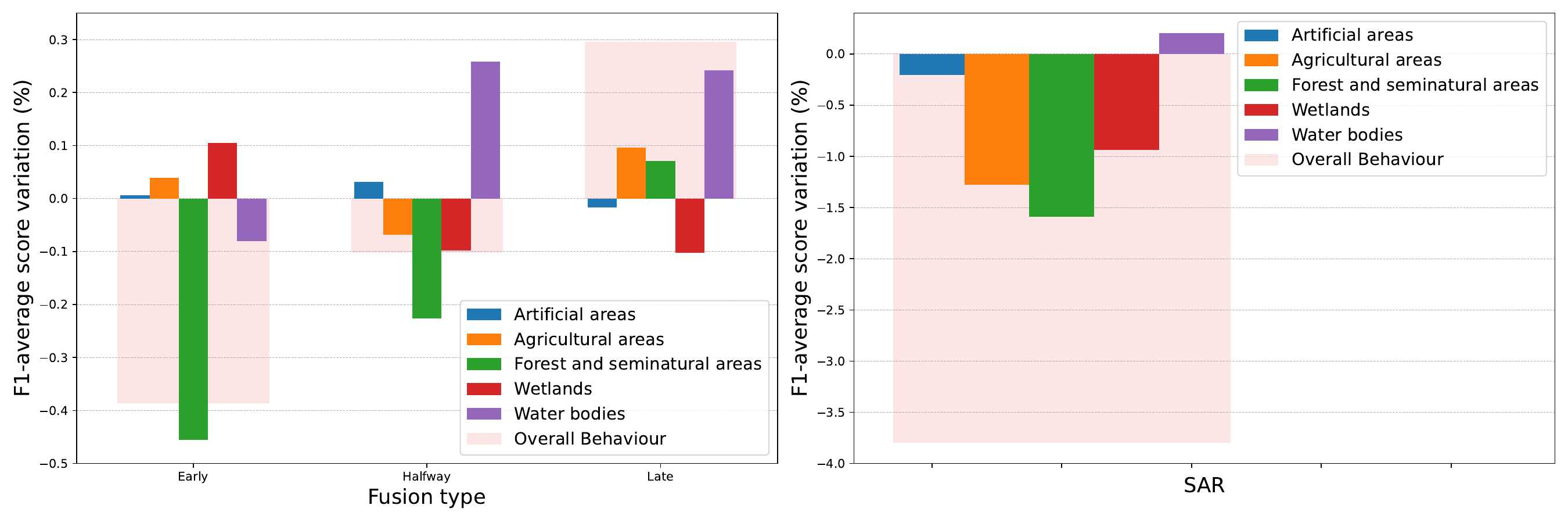}
    \caption{Percental variation in the behaviour of each L1 class based on the F1-average score, comparing results from the three fusion methods and SAR images alone to those from optical images.}
    \label{classimprov}
\end{figure*}

Figure~\ref{classimprov} shows the impact of each fusion method on individual L1 classes, reflected by changes in the F1-average scores. These scores account for dataset imbalance and compare the performance of three fusion methods and SAR images with optical images. The model applied to SAR images is generally less accurate for most L1 classes, with the exception of \textit{Water bodies}.  
Halfway and late fusion make the most of the improvement of SAR images on the \textit{Water bodies}, with even larger improvement with respect to SAR modality only. 
On the other hand, both these fusions are ineffective for the prediction of the \textit{Wetlands} class, where only the early fusion can improve the prediction accuracy. 
The low performance of wetlands classification can be attributed to several factors. First, while SAR data, particularly HH polarization, is effective for classifying more structured areas like swamps and uplands, this polarization is not present in the dataset. Additionally, the class "Swamps" is absent from our dataset, and many of the wetlands that are present, such as fens, bogs, and marshes, exhibit less structured vegetation, leading to misclassification~\cite{adeli2020wetland}. Moreover, fusing SAR with optical data may introduce inconsistencies due to differences in resolution and spectral properties~\cite{sahour2021integrating}.
To enhance classification accuracy, the use of multispectral Sentinel-2 bands has been shown to improve detection, particularly for smaller wetlands~\cite{sahour2021integrating}. Similarly, Digital Elevation Models (DEM) have proven useful in distinguishing wetland classes~\cite{adeli2020wetland}. This aligns with the findings of~\cite{tosato2024sarimproversvqaperformance}, which demonstrated that adding a ratio channel, and thus incorporating more volumetric information, improved wetland classification by 2\%.

In the \textit{Artificial area} class, late fusion succeeds in improving only two L3 classes. In contrast, halfway fusion significantly improves the \textit{Artificial area} class, particularly in three high-frequency classes, whereas early fusion benefits six smaller classes. 
In the case of \textit{Forest and seminatural areas}, halfway fusion and late fusion improve almost the same classes, but late fusion has a major impact.
Similarly, for \textit{Agricultural Areas} both early and late fusion improve the performances, but late fusion improves more frequent classes.
Our results show that different mergers improve different classes, but what remains common is that if it is not high-frequency classes, the overall performance does not improve significantly.

We next consider how these classification results may influence the VQA performance.
When considering the Hamming distance, the best results are obtained with the late fusion. However, the best matching ratio remains the one obtained with optical images alone. This means that using the late fusion yields fewer errors on average. However, these errors are more spread in the different predictions compared to a model using optical data only.

\subsection{End-to-End RSVQA}
In our End-to-End RSVQA model applied to optical images, we notice that using a ResNet-50 instead of ResNet-152 yields an important performance drop, whether the weights are frozen or fine-tuned.

With SAR, performances remain low both when using ResNet-152 and ResNet-50. The strategy of using weights pre-trained on the BEN-MM classification task does not improve the results. 
This may be due to the misalignment between textual features and those extracted from the visual model.

Indeed, textual and visual representations come from different domains. This may cause a lack of direct correspondence between the two representations, complicating the alignment and fusion of information. This misalignment could be improved using more advanced fusion strategies~\cite{chappuis2021find}.

\begin{table*}
    \centering
    \small
    \begin{tabular}{|c|c|c|c|c||c|c|c|c|}
        \hline
        \textbf{} &
        \textbf{Modality} & \textbf{Network} & \textbf{Pre-trained} & \textbf{Frozen?} & \textbf{Yes/No} & \textbf{Land Cover} & \textbf{Overall}& \textbf{Overall} \\
        & & & & & \textbf{Accuracy}& \textbf{Accuracy}& \textbf{Accuracy} & \textbf{$L_{B_{score}}$ ↑} \\
        \hline
         a&OPT & ResNet-152 & ImageNet & Yes & \textbf{80.03\%} & \textbf{20.71\%} & \textbf{69.94\%} & \textbf{17.94\%} \\
        \hline
         b&OPT & ResNet-152 & BEN-MM & Yes & 71.34\% & 13.44\%- & 61.49\%& 9.49\% \\
        \hline
         c&OPT & ResNet-50 & ImageNet & Yes & 72.17\% & 13.91\% & 62.26\%& 10,26\%\\
        \hline
         d&OPT & ResNet-50 & BEN-MM & Yes & 71.52\% & 13.86\% & 62.31\%& 10.31\%\\
        \hline
         e&OPT & ResNet-50 & ImageNet & No & 71.55\% & 12.17\% & 61.46\% & 9.49\%\\
        \hline
        \hline
         f&SAR & ResNet-152 & ImageNet & Yes & 62.66\%	& 15.23\%& 55.09\%& 3.09\%\\
        \hline
         g&SAR & ResNet-152 & BEN-MM & Yes & 61.97\%	&15.26\% & 54.51\%& 2.51\%\\
        \hline
         h&SAR & ResNet-50 & ImageNet & Yes & 71.54\%& 14.24\% & 61.79\%& 9.79\%\\
        \hline
         i&SAR & ResNet-50 & BEN-MM & Yes & 71.21\% & 13.50\% & 61.49\%& 9.49\%\\
        \hline
         j&SAR & ResNet-50 & ImageNet & No & 72.08\%& 14.14\%&62.23\% & 10.23\%\\
         \hline
         %k&Black Images&&&&72.04\% ± 0.92\%& 13.63\% ± 0.50\%&62.10\% ± 0.78\%\\
        %hline
    \end{tabular}
    \caption{End-to-End RSVQA results on the RSVQAxBEN-MM dataset.}
    \label{tab:RSVQAexp}
\end{table*}

\subsection{Prompt-RSVQA}
The results of the Prompt-RSVQA method build on the classification results presented in Section~\ref{classificationdiscussion}, even if improvements in classification performance do not directly translate to equivalent gains in VQA accuracy. %the higher the accuracy, the more classes are predicted and more questions are correctly answered.  
Indeed, an important observation in classification is that the ResNet-50 model shows a 3.7\% difference in MR between optical and SAR predictions. However, this results in a 3.62\% increase in Accuracy.

For both SAR and optical images we can observe that the decrease in the  F1-micro classification scores when increasing the network depth is reflected in the corresponding VQA results. 
It is interesting to notice in Table~\ref{tab:prompt}[c,d] that with ResNet-152 the land cover accuracy obtained with SAR improves by 0.64\%.  However, these questions are under-represented in the dataset, which leads to an inferior overall accuracy compared to the one obtained with ResNet-50.

Regarding the fusion results, we observe a gradual improvement from early fusion, halfway fusion to late fusion. This is coherent with what we observe in the classification results. However, late fusion is the only one that outperforms a pipeline using optical images only, especially in improving the land cover accuracy. This is linked to the match ratio results.

Although these relatively simple fusion strategies already provide better performance, it should be noted that more complex approaches have also been explored. For instance, methods such as VisualBERT~\cite{li2019visualbert} have been applied in remote sensing VQA settings~\cite{siebert2022multi}. However, despite leveraging attention-based fusion mechanisms, the performance reported in that work remains below the level achieved by our simpler architecture. %\textcolor{orange}{(Reviewer 1, comment 1)}

\begin{figure*}[t!]
    \centering
    \includegraphics[width = \textwidth]{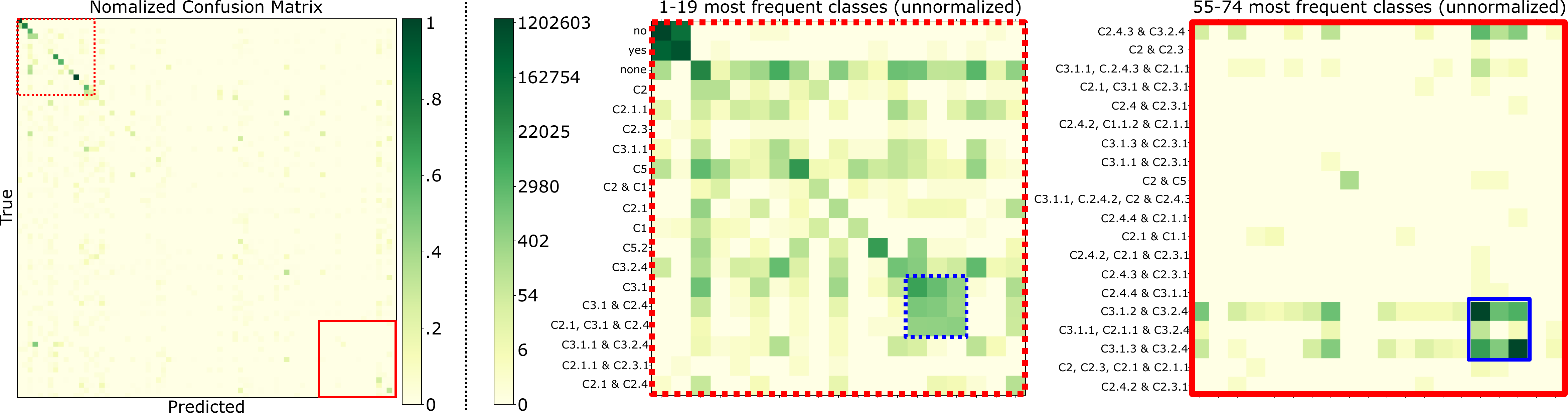}
    \caption{ 
    Confusion matrices of the Prompt-RSVQA model with a ResNet-50 backbone and a late fusion mechanism. On the left-hand side is the normalized confusion matrix. On the right-hand side, two zooms in the logarithm scale are represented. "CX" corresponds to class X in the CLC taxonomy, available \href{https://land.copernicus.eu/content/corine-land-cover-nomenclature-guidelines/html/}{online}~\textsuperscript{1}.
    }
    \label{confusionmatlate}
\end{figure*}

Figure~\ref{confusionmatlate} represents the normalized confusion matrix for the Prompt-RSVQA method with a ResNet-50 backbone and a late fusion mechanism. 
On the left of Figure~\ref{confusionmatlate}, the normalized confusion matrix with the 75 more frequent answers is presented ranked by frequency. 
On the right, two zooms of the confusion matrix are presented. Note that to enrich our interpretation, the zooms are presented in a non-normalized (and in log-scale) version. The top-right zoom highlights the 19 most frequent classes, while the bottom-right zoom displays the classes from the 55th to the 74th most frequent ones.

\begin{figure*}
    \centering
    \includegraphics[width=\textwidth]{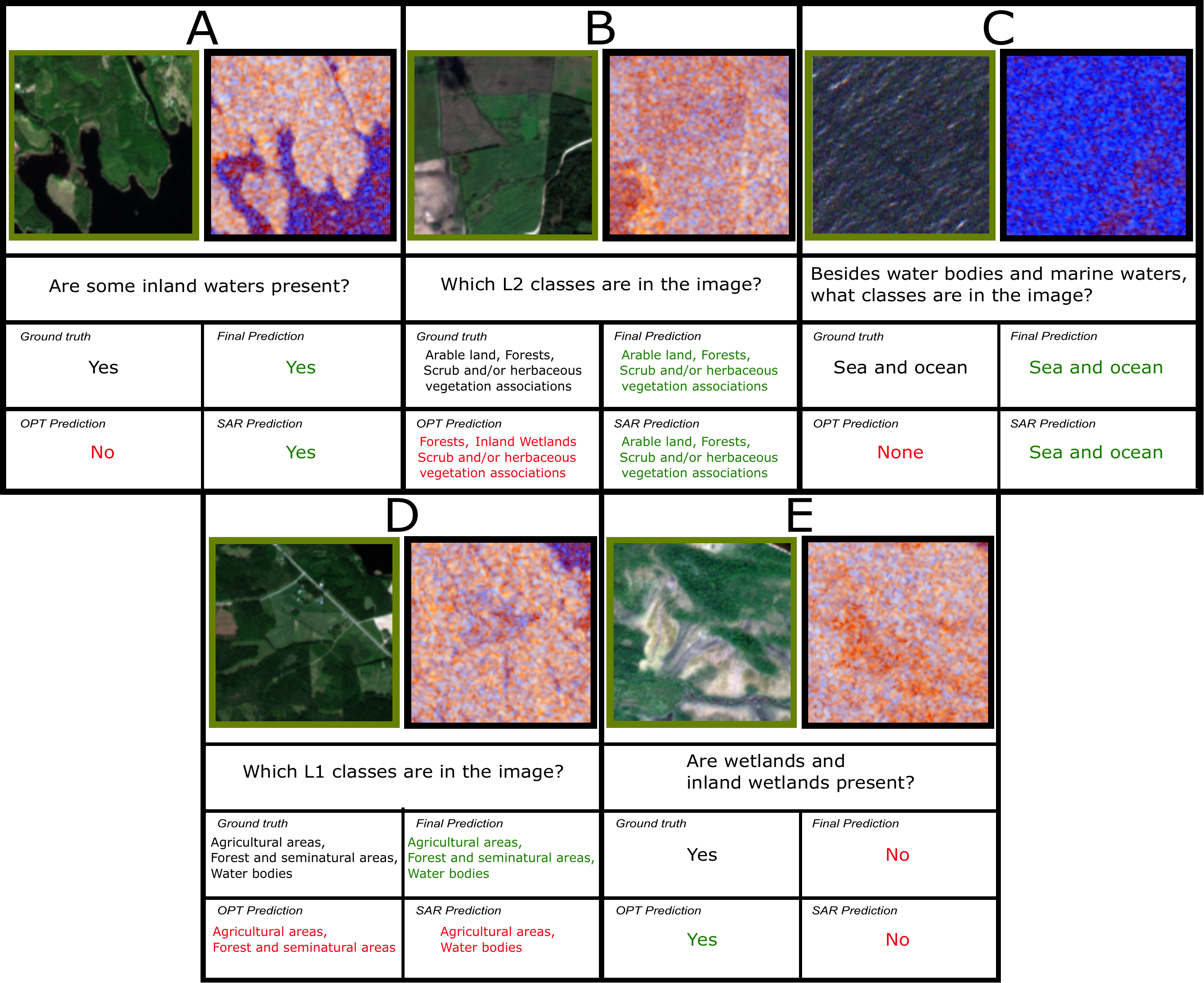}
    \caption{Visual results from the test set. The two modalities, the prediction of OPTICAL and SAR image only and the late fusion prediction are represented.
    }
    \label{fig:resultsexamples}
\end{figure*}

In the normalized confusion matrix, we observe that for the 30 most-frequent answers the trend is relatively correct, with a visible diagonal. 
In the non-normalised confusion matrix on the top right corner, the strong imbalance in the answers clearly appears. Indeed, 82\% of questions of the test set are \textit{yes/no} questions. As seen in Table~\ref{tab:prompt}[g], our model can predict the answer to these questions with good accuracy. 
Another fact we can observe is that the model struggles to distinguish between classes that belong to the same branch of the CORINE Land Cover hierarchy. Two examples of this phenomenon can be illustrated in the blue bounding boxes with answers from the same hierarchy being ordered consecutively in the matrix. 
Visually, we show in Figure~\ref{fig:resultsexamples} some predictions made by Prompt-RSVQA on the test set.

%For the others, we can see that there are still many false positives. \red{add something} it's a decent confusion matrix , model struggle to difff the different level of the CLC map. 
\begin{table*}
    \centering
    \small
    \begin{tabular}{|c|c|c|c|c||c|c|c|c|}
        \hline
        \textbf{} & 
        \textbf{Modality} & \textbf{Network} & \textbf{Pre-trained} & \textbf{Frozen?} & \textbf{Yes/No } & \textbf{Land Cover } & \textbf{Overall} & \textbf{Overall}\\
        & & & & & \textbf{Accuracy}& \textbf{Accuracy} & \textbf{Accuracy} & \textbf{$L_{B_{score}}$ ↑}\\
        \hline
         a&OPT & ResNet-152 & BEN-MM & Yes & 84.55\% & 24.93\% & 73.86\% & 21.86\% \\
        \hline
         b&OPT & ResNet-50 & BEN-MM & Yes & \textbf{86.07\%} & 26.56\% & 75.40\% & 23.40\%  \\
        \hline
        \hline
         c&SAR & ResNet-152 & BEN-MM & Yes & 82.80\% & 21.28\%& 71.77\% & 19.77\%\\
        \hline
         d&SAR & ResNet-50 & BEN-MM & Yes & 82.94\% & 20.64\% & 71.78\% & 19.78\%\\
        \hline
        \hline
         e&Early & ResNet-50 & BEN-MM & Yes & 85.54\% & 26.03\% & 74.88\% & 22.88\%   \\
         & Fusion & & & & & & & \\
        \hline
         f&Halfway & ResNet-50 & BEN-MM & Yes & 85.90\% &	25.91\% & 75.15\% & 23.15\% \\
        & Fusion & & & & & & & \\
        \hline
         g&Late & ResNet-50 & BEN-MM & Yes & \textbf{86.07\%} & \textbf{27.03\%} & \textbf{75.49\%} & \textbf{23.49\%} \\
         & Fusion & & & & & & & \\
         \hline
    \end{tabular}
    \caption{Prompt-RSVQA results on RSVQAxBEN-MM dataset using ResNet with different depth, modality as input and different types of data fusion. }
    \label{tab:prompt}
\end{table*}

\section{Conclusion} \label{conclusion}
\begin{comment}
In this article, we present RSVQAxBEN-MM, a new dataset incorporating SAR images into RSVQA. To our knowledge, this is the first attempt to explore RSVQA for land cover classification using high-resolution SAR imagery, compare its performance with optical images taken under clear conditions and during daylight, and investigate fusion strategies. We extend two pipelines to integrate SAR data and find that the end-to-end RSVQA model performs well only with optical data when using ResNet-152, highlighting its lack of robustness. However, Prompt-RSVQA improves performance with SAR, suggesting that visual features may be too complex for effective interaction with textual features in networks not extensively trained for SAR. Models relying solely on SAR underperform compared to optical data, but late fusion demonstrates that SAR provides valuable information, particularly for water-related classes. 
The effectiveness of fusion methods is dataset-dependent, with late fusion performing best on our imbalanced dataset, while early and halfway fusion benefit underrepresented classes. Future work could focus on creating more balanced RSVQA datasets with SAR, broader question coverage, and improved distributions.
\end{comment}
In this article, we introduce a new dataset including SAR images as a modality for RSVQA: RSVQAxBEN-MM. To the best of our knowledge, our work represents the first attempt to explore RSVQA for land cover classification using high-resolution SAR imagery, compare its performance with optical images taken under clear conditions and during daylight, and investigate fusion strategies. 
Moreover, we extend two pipelines to use the SAR data as input. Our experiments with the End-to-End RSVQA model show that it only performs well with optical data when using ResNet-152 as an image encoder, indicating that it is not a robust model. We show that with Prompt-RSVQA, we can gain in performances with the SAR modality. 
%This suggests that the visual features might be too complex to interact effectively with textual features, especially in networks that are not extensively trained, such as ResNet with ImageNet for other modalities. 
This suggests that visual features may be too complex for effective interaction with textual features in networks that have not been extensively trained for SAR data. Furthermore, a model relying on SAR data alone does not achieve results as high as optical data. However, using late fusion, we notice that SAR can add relevant information, particularly in water-related classes. The choice of the fusion method, however, strongly depends on the dataset used. On our proposed dataset, which is imbalanced across classes, late fusion is the most effective method. This result may not hold for other datasets with a different data distribution. Indeed, while early and halfway fusion methods improve the classification results on more classes, these classes are under-represented in questions present in the dataset.
As such, future work could be focused on developing new RSVQA datasets including SAR as a modality, with better question coverage and with improved data distributions.
\footnotetext[1]{\url{https://land.copernicus.eu/content/corine-land-cover-nomenclature-guidelines/html/}}
\bibliographystyle{IEEEtran}
\bibliography{ref}
\newpage

\section{Biography Section}
%If you have an EPS/PDF photo (graphicx package needed), extra braces are needed around the contents of the optional argument to biography to prevent the LaTeX parser from getting confused when it sees the complicated
% $\backslash${\tt{\includegraphics}} command within an optional argument. (You can create your own custom macro containing the $\backslash${\tt{\includegraphics}} command to make things simpler here.)
 
\vspace{11pt}

%\bf{If you include a photo:}\vspace{-33pt}

\begin{IEEEbiography}[{\includegraphics[width=1in,height=1.25in,clip,keepaspectratio]{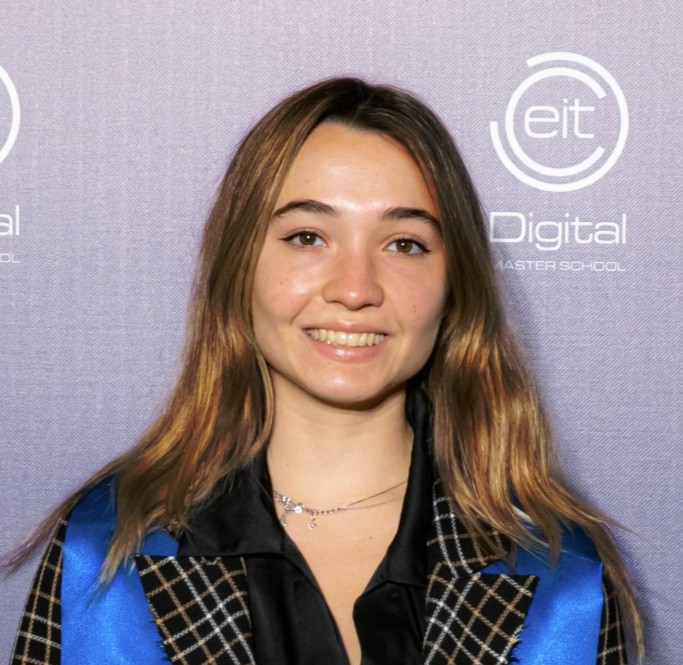}}]{Lucrezia Tosato}
is a PhD student in image processing at Université Paris Cité and the French Aerospace Laboratory (ONERA), France. She holds a Master's degree in Computer Science from Sorbonne University and an MsEng degree from the University of Trento. Her research interests include computer vision, natural language processing, SAR imagery, and data interpretability and explainability.
\end{IEEEbiography}

\begin{IEEEbiography}[{\includegraphics[width=1in,height=1.25in,clip,keepaspectratio]{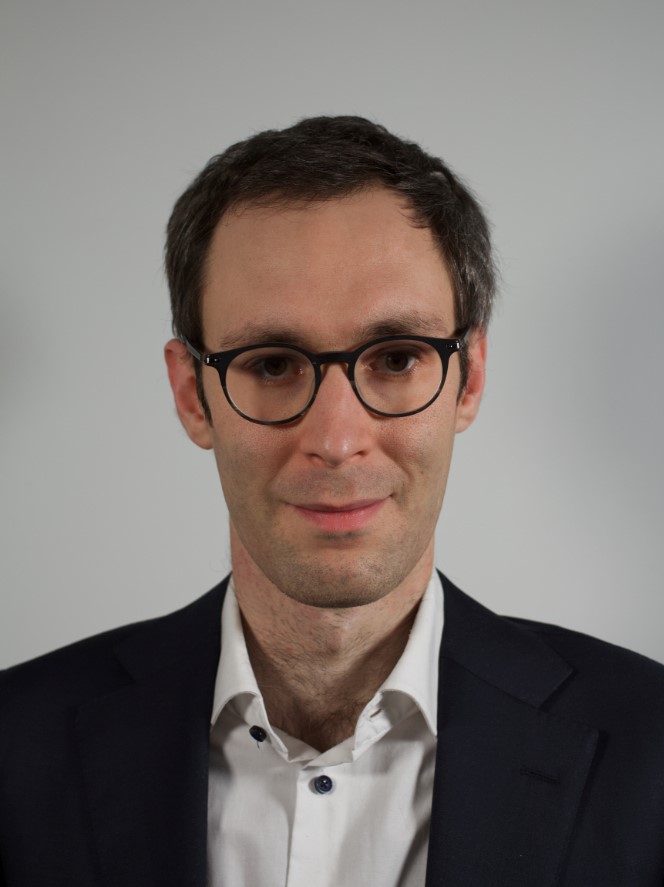}}]{Sylvain Lobry}
is an Assistant Professor at the LIPADE Laboratory, Université de Paris, Paris, France. He received his PhD degree in signal and image processing from Télécom Paris, France, in 2017. From 2017 to 2020, he was a postdoctoral researcher at Wageningen University, the Netherlands. His research interests include image processing and machine learning methods, for example Visual Question Answering, using remote sensing data.
\end{IEEEbiography}

\begin{IEEEbiography}[{\includegraphics[width=1in,height=1.25in,clip,keepaspectratio]{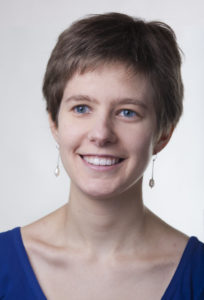}}]{Flora Weissgerber}
Flora Weissgerber is a research director at the French Aerospace Laboratory (ONERA), in the SAPIA team of the Information Processing and Systems department, based in Palaiseau, France. She works on the analysis of muti-modal remote sensing, including SAR images, Optical images and radar altimetry. Her focus on SAR images includes interferometry and change detection.
\end{IEEEbiography}

\begin{IEEEbiography}[{\includegraphics[width=1in,height=1.25in,clip,keepaspectratio]{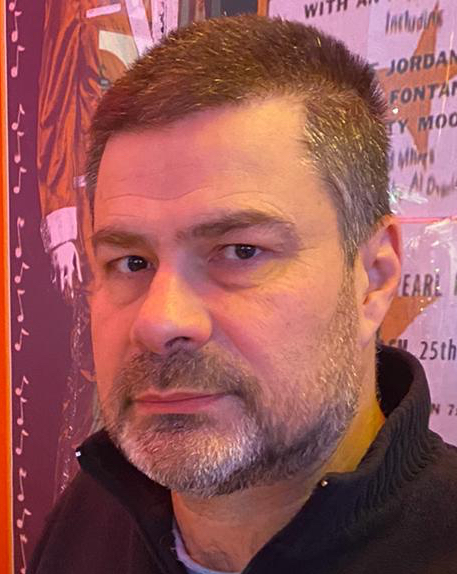}}]{Laurent Wendling}
is a full professor at the LIPADE Laboratory, Université Paris Cité, Paris, France. He received his PhD degree in computer science from Université Paul Sabatier, Toulouse, France, in 1997. His research interests include pattern recognition and computer vision.
\end{IEEEbiography}

\vspace{11pt}

%\bf{If you will not include a photo:}\vspace{-33pt}
%\begin{IEEEbiographynophoto}{John Doe}
%Use $\backslash${\tt{begin\{IEEEbiographynophoto\}}} and the author name as the argument followed by the biography text.
%\end{IEEEbiographynophoto}

\vfill

\end{document}